\documentclass[runningheads]{llncs}

\makeatletter
\newcommand{\@chapapp}{\relax}%
\makeatother

\usepackage[title,toc,titletoc,header]{appendix}

\usepackage[table,xcdraw]{xcolor}

\usepackage{graphicx}
\usepackage{amsmath,amssymb} 
\usepackage{placeins}
\usepackage{color}
\usepackage[width=122mm,left=12mm,paperwidth=146mm,height=193mm,top=12mm,paperheight=217mm]{geometry}

\usepackage[colorlinks=true]{hyperref}
\usepackage{gensymb}
\usepackage{subfigure}
\usepackage{verbatimbox}

\usepackage{tikz}

\usetikzlibrary{external}
\newlength\figureheight
\newlength\figurewidth

\newcommand{\fig}[1]{Figure~\ref{fig:#1}}
\newcommand{\sect}[1]{Section~\ref{sect:#1}}

\newcommand{\eq}[1]{(\ref{eq:#1})}

\newcommand{\Fsub}[1]{\mathbf{F}_{\mbox{\scriptsize #1}}}
\newcommand{\Dsub}[1]{\mathbf{D}_{\mbox{\scriptsize #1}}}
\newcommand{\mssize}[1]{\mbox{\scriptsize #1}}

\newcommand{\suppmat}{the Appendix}

\makeatletter
\renewcommand\p@subfigure{\thefigure}

\makeatother

\definecolor{inodefill}{HTML}{D9EAD3}
\definecolor{inodedraw}{HTML}{B6D7A8}
\definecolor{pnodefill}{HTML}{CFE2F3}
\definecolor{pnodedraw}{HTML}{9FC5E8}
\definecolor{onodefill}{HTML}{E6B8AF}
\definecolor{onodedraw}{HTML}{CC4125}
\definecolor{mnodefill}{HTML}{EAD1DC}
\definecolor{mnodedraw}{HTML}{D5A6BD}
\definecolor{goldfill}{HTML}{EAD7AC}
\definecolor{golddraw}{HTML}{D1AC71}
\definecolor{bluefill}{HTML}{5AB1F2}
\definecolor{bluedraw}{HTML}{0083E5}

\makeatletter
\newcommand{\todo}[1][]{\@latex@warning{TODO #1}}
\makeatother

\begin{document}
\pagestyle{headings}
\mainmatter
\def\ECCV16SubNumber{332}  

\title{DeepWarp: Photorealistic Image\\ Resynthesis for Gaze Manipulation}

\author{Yaroslav Ganin, Daniil Kononenko, Diana Sungatullina, Victor Lempitsky}
\institute{Skolkovo Institute of Science and Technology,\\
\email{\{ganin,daniil.kononenko,d.sungatullina,lempitsky\}@skoltech.ru}
}

\maketitle

\begin{abstract}
In this work, we consider the task of generating highly-realistic images of a given face with a redirected gaze. We treat this problem as a specific instance of conditional image generation and suggest a new deep architecture that can handle this task very well as revealed by numerical comparison with prior art and a user study. Our deep architecture performs coarse-to-fine warping with an additional intensity correction of individual pixels. All these operations are performed in a feed-forward manner, and the parameters associated with different operations are learned jointly in the end-to-end fashion. After learning, the resulting neural network can synthesize images with manipulated gaze, while the redirection angle can be selected arbitrarily from a certain range and provided as an input to the network.
\end{abstract}

\renewcommand{\floatpagefraction}{0.95}
\section{Introduction}


In this work, we consider the task of learning deep architectures that can transform input images into new images in a certain way (deep image resynthesis). Generally, using deep architectures for image generation has become a very active topic of research. While a lot of very interesting results have been reported over recent years and even months, achieving photo-realism beyond the task of synthesizing small patches has proven hard. 

Previously proposed methods for deep resynthesis usually tackle the resynthesis problem in a general form and strive for universality. Here, we take an opposite approach and focus on a very specific image resynthesis problem (gaze manipulation) that has a long history in the computer vision community~\cite{Okada94,Yang02,Yip03,Criminisi03,Jones09,Wolf10,Kuster12,Giger14,Kononenko15} and some important real-life applications. We show that by restricting the scope of the method and exploiting the specifics of the task, we are indeed able to train deep architectures that handle gaze manipulation well and can synthesize output images of high realism (\fig{teaser}).

\newcommand{\teasercol}[1]{%
\begin{tikzpicture}
\foreach \y[evaluate=\y as \emy using int(4 - \y)] in {0,...,4} {
  \node [anchor=north west,inner sep=0] at (0, \y * 1.1cm) {\pgfimage[height=1.1cm]{./figures/images/teaser/#1_00\emy.png}};
}
\end{tikzpicture}%
}
\tikzexternaldisable
\begin{figure*}
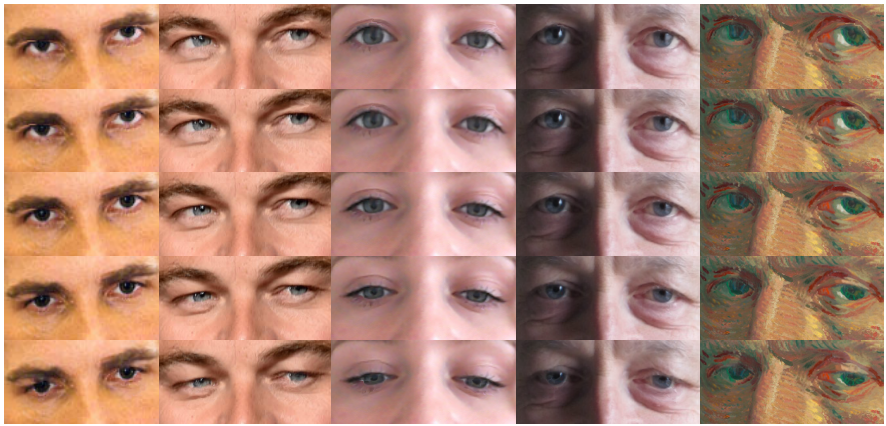

\centering
\setlength{\tabcolsep}{0pt}
\renewcommand{\arraystretch}{0}
\begin{tabular}{ccccc}
\teasercol{bale_ms3_palette} &
\teasercol{leo_ms3_palette} &
\teasercol{test1_ms3_palette} &
\teasercol{geoff_ms3_palette} &
\teasercol{van_ms3_palette}
\end{tabular}
\caption{Gaze redirection with our model trained for vertical gaze redirection. The model takes an input image (middle row) and the desired redirection angle (here varying between -15 and +15 degrees) and re-synthesize the new image with the new gaze direction. Note the preservation of fine details including specular highlights in the resynthesized images. }
\label{fig:teaser}
\end{figure*}
\tikzexternalenable

Generally, few image parts can have such a dramatic effect on the perception of an image like regions depicting eyes of a person in this image. Humans (and even non-humans \cite{Wallis15}) can infer a lot of information about of the owner of the eyes, her intent, her mood, and the world around her, from the appearance of the eyes and, in particular, from the direction of the gaze. Generally, the role of gaze in human communication is long known to be very high~\cite{Kleinke86}.

In some important scenarios, there is a need to digitally alter the appearance of eyes in a way that changes the apparent direction of the gaze. These scenarios include gaze correction in video-conferencing, as the intent and the attitude of a person engaged in a videochat is distorted by the displacement between the face on her screen and the webcamera (e.g.\ while the intent might be to gaze into the eyes of the other person, the apparent gaze direction in a transmitted frame will be downwards). Another common scenario that needs gaze redirection is  ``talking head''-type videos, where a speaker reads the text appearing alongside the camera but it is desirable to redirect her gaze into the camera. One more example includes editing of photos (e.g.\ group photos) and movies (e.g.\ during postproduction) in order to make gaze direction consistent with the ideas of the photographer or the movie director. 

All of these scenarios put very high demands on the realism of the result of the digital alteration, and some of them also require  real-time or near real-time operation. To meet these challenges, we develop a new deep feed-forward architecture that combines several principles of operation (coarse-to-fine processing, image warping, intensity correction). The architecture is trained end-to-end in a supervised way using a specially collected dataset that depicts the change of the appearance under gaze redirection in real life. 

Qualitative and quantitative evaluation demonstrate that our deep architecture can synthesize very high-quality eye images, as required by the nature of the applications, and does so at several frames per second. Compared to several recent methods for deep image synthesis, the output of our method contains larger amount of fine details (comparable to the amount in the input image). The quality of the results also compares favorably with the results of a random forest-based gaze redirection method~\cite{Kononenko15}. Our approach has thus both practical importance in the application scenarios outlined above, and also contributes to an actively-developing field of image generation with deep models.

\section{Related work}

\subsubsection{Deep learning and image synthesis.}
Image synthesis using neural networks is receiving growing attention~\cite{Mahendran15,Dosovitskiy15,Goodfellow14,Denton15,Gatys15,Gregor15}. More related to our work are methods that learn to transform input images in certain ways~\cite{Kulkarni15,Ghodrati15,Reed15}. These methods proceed by learning internal compact representations of images using encoder-decoder (autoencoder) architectures, and then transforming images by changing their internal representation in a certain way that can be trained from examples. We have conducted numerous experiments following this approach combining standard autoencoders with several ideas that have reported to improve the result (convolutional and up-convolutional layers \cite{Zeiler14,Dosovitskiy15}, adversarial loss~\cite{Goodfellow14}, variational autoencoders~\cite{Kingma14}). However, despite our efforts (see \suppmat), we have found that for large enough image resolution, the outputs of the network lacked high-frequency details and were biased towards typical mean of the training data (``regression-to-mean'' effect). This is consistent with the results demonstrated in \cite{Kulkarni15,Ghodrati15,Reed15} that also exhibit noticeable blurring.

Compared to \cite{Kulkarni15,Ghodrati15,Reed15}, our approach can learn to perform a restricted set of image transformations. However, the perceptual quality and, in particular, the amount of high-frequency details is considerably better in the case of our method due to the fact that we deliberately avoid any input data compression within the processing pipeline. This is crucial for the class of applications that we consider.

Finally, the idea of spatial warping that lies in the core of the proposed system has been previously suggested in \cite{Jaderberg15}. In relation to \cite{Jaderberg15}, parts of our architecture can be seen as spatial transformers with the localization network directly predicting a sampling grid instead of low-dimensional transformation parameters.

\subsubsection{Gaze manipulation.}
An early work on monocular gaze manipulation~\cite{Wolf10} did not use machine learning, but relied on pre-recording a number of potential eye replacements to be copy-pasted at test time. The idea of gaze redirection using supervised learning was suggested in \cite{Kononenko15}, which also used warping fields that in their case were predicted by machine learning. Compared to their method, we use deep convolutional network as a predictor, which allows us to achieve better result quality. Furthermore, while random forests in~\cite{Kononenko15} are trained for a specific angle of gaze redirection, our architecture allows the redirection angle to be specified as an input, and to change continuously in a certain range. Most practical applications discussed above require such flexibility. Finally, the realism of our results is boosted by the lightness adjustment module, which has no counterpart in the approach of \cite{Kononenko15}.

Less related to our approach are methods that aim to solve the gaze problem in videoconferencing via synthesizing 3D rotated views of either the entire scene \cite{Okada94,Criminisi03,Yang02} or of the face (that is subsequently blended into the unrotated head) \cite{Kuster12,Giger14}. Out of this works only \cite{Giger14} works in a monocular setting without relying on extra imaging hardware. The general problem with the novel view synthesis is how to fill disoccluded regions. In cases when the 3D rotated face is blended into the image of the unrotated head \cite{Kuster12,Giger14}, there is also a danger of distorting head proportions characteristic to a person.

\section{The model}
\label{sect:model}

%
%
\newcommand{\nnfeaturemap}[5]{%
  \ifthenelse{\equal{#4}{}}{%
    \draw[#5] (#2,0,#3) -- (-#2,0,#3) -- (-#2,0,-#3) -- (#2,0,-#3) -- cycle;
  }{}
  \coordinate (#1@tl) at (#2,0,#3);
  \coordinate (#1@tr) at (-#2,0,#3);
  \coordinate (#1@br) at (-#2,0,-#3);
  \coordinate (#1@bl) at (#2,0,-#3);
  \coordinate (#1@c) at (0,0,0);
}

%
%
\newcommand{\nnimage}[6]{%
  \ifthenelse{\equal{#4}{}}{%
    \begin{scope}[canvas is xz plane at y=0]
      \node[transform shape] (a) {\includegraphics[height=2850pt,width=2850pt]{#6}};
    \end{scope}
    \draw[#5] (#2,0,#3) -- (-#2,0,#3) -- (-#2,0,-#3) -- (#2,0,-#3) -- cycle;
  }{}
  \coordinate (#1@tl) at (#2,0,#3);
  \coordinate (#1@tr) at (-#2,0,#3);
  \coordinate (#1@br) at (-#2,0,-#3);
  \coordinate (#1@bl) at (#2,0,-#3);
  \coordinate (#1@c) at (0,0,0);
}

%
%
\newcommand{\nnanchors}[3]{%
  \coordinate (#1@1) at (31,0,-6);
  \coordinate (#1@2) at (10,0,5);
  \coordinate (#1@3) at (-8,0,4);
  \coordinate (#1@4) at (-28,0,1);
  \coordinate (#1@5) at (-14,0,-14);
  \coordinate (#1@6) at (10,0,-14);
  \coordinate (#1@7) at (1,0,-4);
  
	\draw[#2] (#1@1) circle (#3);
  \draw[#2] (#1@2) circle (#3);
  \draw[#2] (#1@3) circle (#3);
  \draw[#2] (#1@4) circle (#3);
  \draw[#2] (#1@5) circle (#3);
  \draw[#2] (#1@6) circle (#3);
  \draw[#2] (#1@7) circle (#3);
}

%
%
\def\dispsx{{%
0,0,0,0,0,%
-2,-2,0,2,2,%
0,-2,0,2,0,%
-2,-2,0,2,2,%
0,0,0,0,0}}
\def\dispsy{{%
0,0,0,0,0,%
-4,-6,-10,-6,-4,%
0,-3,-8,-3,0,%
4,4,6,4,4,%
0,0,0,0,0}}
\newcommand{\nnflow}[5]{%
  \ifthenelse{\equal{#4}{}}{%
    \draw[#5] (#2,0,#3) -- (-#2,0,#3) -- (-#2,0,-#3) -- (#2,0,-#3) -- cycle;
    \foreach \x in {1,...,4} {
    	\foreach \y in {1,...,4} {
				\draw[#5] (-#2,0,-#3+20*\y) -- (#2,0,-#3+20*\y);
				\draw[#5] (-#2+20*\x,0,-#3) -- (-#2+20*\x,0,#3);
			}
    }
    \foreach \x in {0,...,4} {
    	\foreach \y in {0,...,4} {
				\pgfmathparse{5*\y+\x}
				\pgfmathsetmacro{\dispx}{2*\dispsx[\pgfmathresult]}
				\pgfmathsetmacro{\dispy}{2*\dispsy[\pgfmathresult]}
				\pgfmathparse{\dispx+\dispy}
				\ifthenelse{\lengthtest{\pgfmathresult pt = 0 pt}}{%
				  \fill[black] (#2 - 10 - 20*\x, 0, #3 - 10 - 20*\y) circle (1pt);
				}{%
				  \draw[stealth-,line width=1pt,black] (#2 - 10 - 20*\x - 0.5*\dispx, 0, #3 - 10 - 20*\y - 0.5*\dispy) -- +(\dispx, 0, \dispy);
				}
			}
		}
  }{}
  \coordinate (#1@tl) at (#2,0,#3);
  \coordinate (#1@tr) at (-#2,0,#3);
  \coordinate (#1@br) at (-#2,0,-#3);
  \coordinate (#1@bl) at (#2,0,-#3);
  \coordinate (#1@c) at (0,0,0);
}

\tikzset{
  double arrow/.style args={#1 colored by #2 and #3}{
    -{Triangle[length=20pt,width=40pt]},line width=#1,#2, 
    postaction={draw,-{Triangle[length=20pt-1.414pt-1pt,width=40pt-2*1.414pt-2pt]},#3,line width=(#1 - 2pt),
                shorten <= 1pt,shorten >= 1.414pt}, 
  }
}

\newcommand{\nnarrow}[4]{%
  \draw[double arrow=20pt colored by #2 and #1] #3;
  \ifthenelse{\equal{#4}{}}{%
  }{%
    \path[postaction={decorate,decoration={raise=-1ex,text along path,text align={center, right indent=10pt},text={|\ttfamily\large|#4}}}] #3;
  }
}

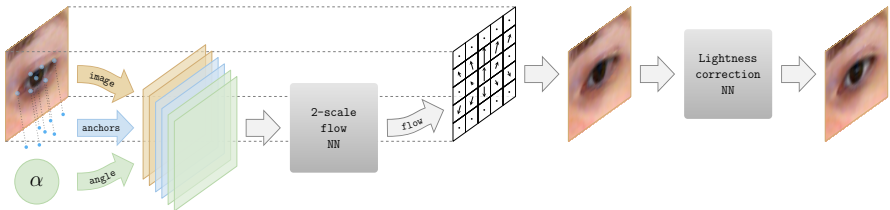
\begin{figure*}[t!]
\centering
\scalebox{0.335}{\begin{tikzpicture}[%
	x={($ (-0.71pt, -0.5pt) $)},y={($ (0.71pt, -0.71pt) $)},z={($ (0pt, 1pt) $)},
  fnode/.style={
    align=center,
    rectangle,minimum height=30pt,minimum width=60pt,rounded corners=2pt,
    inner sep=2pt,
    line width=1pt,
    fill=pnodefill,draw=pnodedraw,
    font=\ttfamily\Large},
  dnode/.style={
    fnode,trapezium,trapezium left angle=70, trapezium right angle=110,trapezium stretches},
  inode/.style={
    dnode,fill=inodefill,draw=inodedraw},
  blackbox/.style={
	  rounded corners,
	  minimum size=100pt,
	  top color=black!10,
	  bottom color=black!30,
	  font=\ttfamily\Large}]

\begin{scope}[shift={(0,0)}]
  \begin{pgfonlayer}{background}
    \nnimage{image}{50}{50}{}{draw=golddraw,line join=round,line width=1pt}{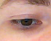}
  \end{pgfonlayer}
  \nnanchors{anchors1}{fill=bluefill!50!white,draw=bluedraw!50!white,opacity=0.6}{2pt}
\end{scope}

\node (image_node) [anchor=south,inner sep=0pt,fit=(image@tr) (image@bl)] {};
\node (angle_node) [circle,minimum size=50pt,fill=inodefill,draw=inodedraw,line width=1pt,anchor=center] at ($ (image_node.center) - (0, 120pt) $) {\Huge$\alpha$};

\begin{scope}[shift={(10pt,-60pt)}]
  \nnanchors{anchors2}{fill=bluefill,draw=white,draw opacity=0.8,line width=2pt}{3pt}
\end{scope}
\begin{pgfonlayer}{background}
  \tikzstyle{anchorline}=[line width=1pt,shorten <= 2pt,dotted,draw=black!80,opacity=0.5]
  \draw[anchorline] (anchors1@1) -- (anchors2@1);
  \draw[anchorline] (anchors1@2) -- (anchors2@2);
  \draw[anchorline] (anchors1@3) -- (anchors2@3);
  \draw[anchorline] (anchors1@4) -- (anchors2@4);
  \draw[anchorline] (anchors1@5) -- (anchors2@5);
  \draw[anchorline] (anchors1@6) -- (anchors2@6);
  \draw[anchorline] (anchors1@7) -- (anchors2@7);
\end{pgfonlayer}

\nnarrow{goldfill}{golddraw}{($ (image_node.east) + (10pt, 0pt) $) .. controls +(0:30pt) and +(160:35pt)  .. +(65pt,-20pt)}{image}
\nnarrow{pnodefill}{pnodedraw}{($ (image_node.east) + (10pt, -60pt) $) -- +(65pt,0)}{anchors}
\nnarrow{inodefill}{inodedraw}{($ (image_node.east) + (10pt, -120pt) $) .. controls +(0:30pt) and +(200:35pt)  .. +(65pt,20pt)}{angle}

\begin{scope}[shift={($ (image_node.east) + (85pt, -60pt) + 50*(0.71pt,0) + 25*(0.71pt,0) $)},local bounding box=input_maps_node]
	\def\fillcolors{{"goldfill","pnodefill","inodefill"}}
	\def\drawcolors{{"golddraw","pnodedraw","inodedraw"}}
	\foreach \y/\c in {-25/0, -15/0, -5/1, 5/1, 15/2, 25/2} {
	  \begin{scope}[shift={(0,\y * 1pt,0)}]
	  	\pgfmathsetmacro{\fillcolor}{\fillcolors[\c]}
			\pgfmathsetmacro{\drawcolor}{\drawcolors[\c]}
		  \nnfeaturemap{dummy}{50}{50}{}{fill=\fillcolor,draw=\drawcolor,fill opacity=0.7,line join=round,line width=1pt}
		\end{scope}
	}
\end{scope}

\nnarrow{black!5}{black!50}{($ (input_maps_node.east) + (10pt, 0) $) -- +(40pt,0)}{}
\node[blackbox,align=center,right=60pt of input_maps_node] (2scale_nn_node) {2-scale\\flow\\NN};

\nnarrow{black!5}{black!50}{($ (2scale_nn_node.east) + (10pt, 0) $) .. controls +(0:30pt) and +(200:35pt)  .. +(65pt,20pt)}{flow}
\begin{pgfonlayer}{background}
  \begin{scope}[shift={($ (2scale_nn_node.east) + (85pt, 60pt) + 50*(0.71pt,0) $)},local bounding box=flow_node]
  	\nnfeaturemap{flow}{50}{50}{nodraw}{line join=round,line width=1pt}
  \end{scope}
  \draw[line width=1pt,dashed,draw=black!50] (image@tl) -- (flow@tl);
  \draw[line width=1pt,dashed,draw=black!50] (image@tr) -- (flow@tr);
  \draw[line width=1pt,dashed,draw=black!50] (image@br) -- (flow@br);
  \draw[line width=1pt,dashed,draw=black!50] (image@bl) -- (flow@bl);
\end{pgfonlayer}
\begin{scope}[shift={($ (2scale_nn_node.east) + (85pt, 60pt) + 50*(0.71pt,0) $)},local bounding box=flow_node]
	\nnflow{flow}{50}{50}{}{fill=white,fill opacity=0.8,draw=black,line join=round,line width=1pt}
\end{scope}

\nnarrow{black!5}{black!50}{($ (flow_node.east) + (10pt, 0) $) -- +(40pt,0)}{}
\begin{scope}[shift={($ (flow_node.east) + (60pt, 0) + 50*(0.71pt,0) $)},local bounding box=pre_output_node]
	\nnimage{pre_output}{50}{50}{}{draw=golddraw,line join=round,line width=1pt}{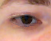}
\end{scope}

\nnarrow{black!5}{black!50}{($ (pre_output_node.east) + (10pt, 0) $) -- +(40pt,0)}{}
\node[blackbox,align=center,right=60pt of pre_output_node] (lcm_node) {Lightness\\correction\\NN};

\nnarrow{black!5}{black!50}{($ (lcm_node.east) + (10pt, 0) $) -- +(40pt,0)}{}
\begin{scope}[shift={($ (lcm_node.east) + (60pt, 0) + 50*(0.71pt,0) $)},local bounding box=output_node]
	\nnimage{output}{50}{50}{}{draw=golddraw,line join=round,line width=1pt}{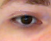}
\end{scope}

\end{tikzpicture}}
\caption{The {\bf proposed system} takes an input eye region, feature points (\texttt{anchors}) as well as a correction \texttt{angle} $ \alpha $ and sends them to the multi-scale neural network (see \sect{methods:warping}) predicting a \texttt{flow} field. The flow field is then applied to the input image to produce an image of a redirected eye. Finally, the output is enhanced by processing with the lightness correction neural network (see \sect{methods:palette}).}
\label{fig:pipeline}
\end{figure*}

In this section, we discuss the architecture of our deep model for re-synthesis. The model is trained on pairs of images corresponding to eye appearance before and after the redirection. The redirection angle serves as an additional input parameter that is provided both during training and at test time.

As in \cite{Kononenko15}, the bulk of gaze redirection is accomplished via warping the input image (\fig{pipeline}). The task of the network is therefore the prediction of the warping field. This field is predicted in two stages in a coarse-to-fine manner, where the decisions at the fine scale are being informed by the result of the coarse stage. Beyond coarse-to-fine warping, the photorealism of the result is improved by performing pixel-wise correction of the brightness where the amount of correction is again predicted by the network. All operations outlined above are implemented in a single feed-forward architecture and are trained jointly end-to-end.

We now provide more details on each stages of the procedure, starting with more detailed description of the data used to train the architecture.

\subsection{Data preparation}
\label{sect:methods:localization}
At training time, our dataset allows us to mine pairs of images containing eyes of the same person looking in two different directions separated by a known angle $ \alpha $. The head pose, the lighting, and all other nuisance parameters are (approximately) the same between the two images in the pair.  Following \cite{Kononenko15} (with some modifications), we extract the image parts around each of the eye and resize them to characteristic scale. For simplicity of explanation, let us assume that we need to handle left eyes only (the right eyes can be handled at training and at test times via mirroring).

To perform the extraction, we employ an external face alignment library \cite{Xiong13} producing, among other things, $ N = 7 $ feature points $ \{ (x_i^{\mssize{anchor}}, y_i^{\mssize{anchor}} ) \, | \, i = 1, \ldots, N  \} $ for the eye (six points along the edge and also the pupil center). Next, we compute a tight axis-aligned bounding box $ \mathcal{B}^{\prime} $ of the points in the \textit{input} image. We enlarge $ \mathcal{B}^{\prime} $ to the final bounding-box $ \mathcal{B} $ using a characteristic radius $ R $ that equals the distance between the corners of an eye. The size of $ \mathcal{B} $ is set to $ 0.8 R \times 1.0 R $. We then cut out the interior of the estimated box from the input image, and also from the output image of the pair (using exactly the same bounding box coordinates). Both images are then rescaled to a fixed size ($ W \times H = 51 \times 41 $ in our experiments). The resulting image pair serves as a training example for the learning procedure (\fig{dataset_collection}-Right). 

\subsection{Warping modules}
\label{sect:methods:warping}

Each of the two warping modules takes as an input the image, the position of the feature points, and the redirection angle. All inputs are expressed as maps as discussed below, and the architecture of the warping modules is thus ``fully-convolutional'', including several convolutional layers interleaved with Batch Normalization layers \cite{Ioffe15} and ReLU non-linearities (the actual configuration is shown in \suppmat). To preserve the resolution of the input image, we use `same'-mode convolutions (with zero padding), set all strides to one, and avoid using  max-pooling. 

\textbf{Coarse warping.} The last convolutional layer of the first (half-scale) warping module produces a pixel-flow field (a two-channel map), which is then upsampled $ \Dsub{coarse}(I, \alpha) $ and applied to warp the input image by means of a bilinear sampler $ \mathbf{S} $ \cite{Jaderberg15,Oquab15} that finds the \textit{coarse estimate}: 
\begin{equation} \label{eq:sampling}
  O_{\mssize{coarse}} = \mathbf{S} \left( I, \Dsub{coarse}(I, \alpha) \right) \, .
\end{equation}
Here, the sampling procedure $S$ samples the pixels of $O_{\mssize{coarse}}$ at pixels determined by the flow field: 
\begin{equation}
O_{\mssize{coarse}}(x,y,c) = I\{x+\Dsub{coarse}(I, \alpha)(x,y,1),y+\Dsub{coarse}(I, \alpha)(x,y,2),c\}\,,    
\end{equation}where $c$ corresponds to a color channel (R,G, or B), and the curly brackets correspond to bilinear interpolation of $I(\cdot,\cdot,c)$ at a real-valued position. The sampling procedure \eq{sampling} is piecewise differentiable \cite{Jaderberg15}.

\textbf{Fine warping.} In the fine warping module, the rough image estimate $O_{\mssize{coarse}}$ and the upsampled low-resolution flow $\Dsub{coarse}(I, \alpha)$ are  concatenated with the input data (the image, the angle encoding, and the feature point encoding) at the original scale and sent to the $ 1\times $-scale network which predicts another two-channel flow $\Dsub{res}$ that amends the half-scale pixel-flow (additively \cite{He15}):
\begin{equation}
  \mathbf{D}(I, \alpha) = \Dsub{coarse}(I, \alpha) + \Dsub{res}(I, \alpha, O_{\mssize{coarse}}, \Dsub{coarse}(I, \alpha)) \, ,
\end{equation}
The amended flow is used to obtain the final output (again, via bilinear sampler):
\begin{equation}
  O = \mathbf{S} \left( I, \mathbf{D}(I, \alpha) \right) \, .  
\end{equation}

The purpose of coarse-to-fine processing is two-fold. The half-scale (coarse) module effectively increases the receptive field of the model resulting in a flow that moves larger structures in a more coherent way. Secondly, the coarse module gives a rough estimate of how a redirected eye would look like. This is useful for locating problematic regions which can only be fixed by a neural network operating at a finer scale. 

\subsection{Input encoding}

As discussed above, alongside the raw input image, the warping modules also receive the information about the desired redirection angle and feature points also encoded as image-sized feature maps.

\textbf{Embedding the angle.} Similarly to \cite{Ghodrati15}, we treat the correction angle as an attribute and embed it into a higher dimensional space using a multi-layer perceptron $ \Fsub{angle} (\alpha) $ with ReLU non-linearities. The precise architecture is \texttt{FC(16)} $ \rightarrow $ \texttt{ReLU} $ \rightarrow $ \texttt{FC(16)} $ \rightarrow $ \texttt{ReLU}. Unlike \cite{Ghodrati15}, we do not output separate features for each spatial location but rather opt for a single position-independent $ 16 $-dimensional vector. The vector is then expressed as $16$ constant maps that are concatenated into the input map stack. During learning, the embedding of the angle parameter is also updated by backpropagation.

\textbf{Embedding the feature points.} Although in theory a convolutional neural network of an appropriate architecture should be able to extract necessary features from the raw input pixels, we found it beneficial to further augment 3 color channels with additional 14 feature maps containing information about the eye anchor points. 

In order to get the anchor maps, for each previously obtained feature point located at $ (x_i^{\mssize{anchor}}, y_i^{\mssize{anchor}}) $, we compute a pair of maps:
\begin{equation}
  \begin{aligned}
  \Delta_x^i[x, y] = x - x_i^{\mssize{anchor}} \, , \\
  \Delta_y^i[x, y] = y - y_i^{\mssize{anchor}} \, , 
  \end{aligned}
  \quad \forall (x, y) \in \{ 0, \ldots, W \} \times \{ 0, \ldots, H \} \, , 
\end{equation}
where $ W, H $ are width and height of the input image respectively. The embedding give the network ``local'' access to similar features as used by decision trees in \cite{Kononenko15}.

Ultimately, the input map stack consists of 33 maps (RGB \textbf{+} 16 angle embedding maps \textbf{+} 14 feature point embedding maps).

\subsection{Lightness Correction Module}
\label{sect:methods:palette}

\newcommand{\paletteimages}[1]{%
\includegraphics{./figures/images/palette_showcase/in_#1.png} &
\includegraphics{./figures/images/palette_showcase/ms3_out_#1.png} &
\includegraphics{./figures/images/palette_showcase/ms3_palette_out_#1.png} &
\includegraphics{./figures/images/palette_showcase/weights_#1.png} &
\includegraphics{./figures/images/palette_showcase/gt_#1.png} \\%
}
\tikzexternaldisable
\begin{figure*}[t!]
\centering
  \setlength{\tabcolsep}{0pt}
  \renewcommand{\arraystretch}{0}
  \begin{tabular}{ccccc}
  Input & CFW & + LCM & Mask & GT \\
  \noalign{\vskip 2pt} \paletteimages{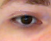}
  \paletteimages{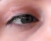}
  \paletteimages{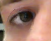}
  \end{tabular}
\caption{Visualization of three challenging redirection cases where \textbf{Lightness Correction Module} helps considerably compared to the system based solely on coarse-to-fine warping (CFW) which is having difficulties with expanding the area to the left of the iris. The `Mask' column shows the soft mask corresponding to parts where lightness is increased. Lightness correction fixes problems with inpainting disoccluded eye-white, and what is more emphasizes the specular highlight increasing the perceived realism of the result.}
\label{fig:palette_examples}
\end{figure*}
\tikzexternalenable

While the bulk of appearance changes associated with gaze redirection can be modeled using warping, some subtle but important transformations are more photometric than geometric in nature and require a more general transformation. In addition, the warping approach can struggle to fill in disoccluded areas in some cases. 

To increase the generality of the transformation that can be handled by our architecture, we add the final lightness adjustment module (see \fig{pipeline}). The module takes as input the features computed within the coarse warping and fine warping modules (specifically, the activations of the third convolutional layer), as well as the image produced by the fine warping module. The output of the module is a single map $M$ of the same size as the output image that  is used to modify the brightness of the output $O$ using a simple element-wise transform: 
\begin{equation} \label{eq:LCM}
   O_\text{final} (x,y,c) = O(x,y,c) \cdot (1-M(x,y)) + M(x,y) \, ,
\end{equation}
assuming that the brightness in each channel is encoded between zero and one. The resulting pixel colors can thus be regarded as blends between the colors of the warped pixels and the white color. The actual architecture for the lightness correction module in our experiments is shown in \suppmat.

This idea can be, of course, generalized further to a larger number of colors in the \textit{palette} for admixing, while these colors can be defined either manually or made dataset-dependent or even image-dependent. Our initial experiments along these directions, however, have not brought consistent improvement in photorealism in the case of the gaze redirection task. 

\section{Experiments}

\subsection{Dataset}

\begin{figure*}
\centering
\includegraphics[width=\textwidth]{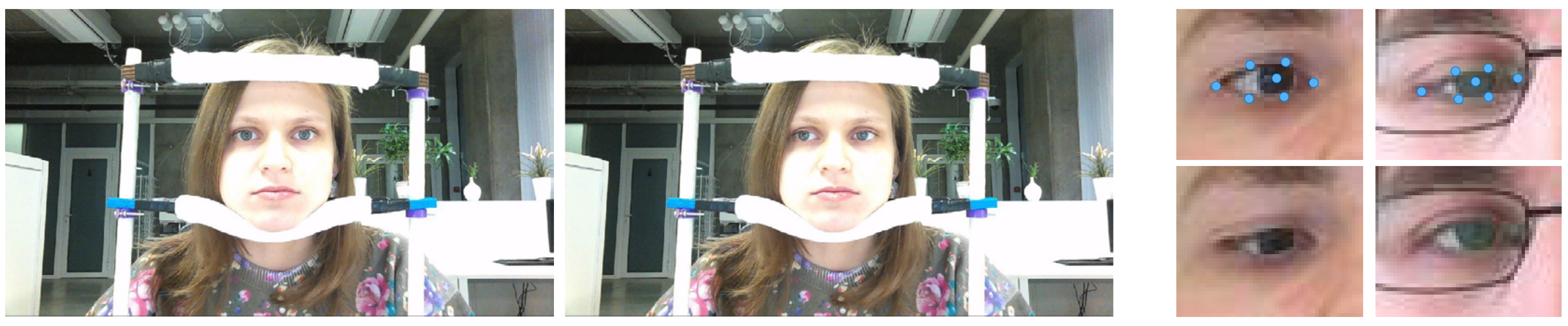}
\caption{Left -- dataset collection process. Right -- examples of two training pairs (input image with superimposed feature points on top, output image in the bottom).}
\label{fig:dataset_collection}
\end{figure*}

There are no publicly available datasets suitable for the purpose of the gaze correction task with continuously varying redirection  angle. Therefore, we collect our own dataset (\fig{dataset_collection}). To minimize head movement, a person places her head on a special stand and follows with her gaze a moving point on the screen in front of the stand. While the point is moving, we record several images with eyes looking in different fixed directions (about $200$ for one video sequence) using a webcam mounted in the middle of the screen. For each person we record $2-10$ sequences, changing the head pose and light conditions between different sequences. Training pairs are collected, taking two images with different gaze directions from one sequence. We manually exclude bad shots, where a person is blinking or where she is not changing gaze direction monotonically as anticipated. Most of the experiments were done on the dataset of $33$ persons and $98$ sequences. Unless noted otherwise, we train the model for vertical gaze redirection in the range between $ -30\degree $ and $ 30\degree $.

\subsection{Training procedure}

The model was trained end-to-end on $128$-sized batches using Adam optimizer \cite{Kingma14}. We used a regular $ \ell_2 $-distance between the synthesized output $O_\text{output}$ and the ground-truth $O_\text{gt}$ as the objective function. We tried to improve over this simple baseline in several ways. First, we tried to put emphasis on the actual eye region (not the rectangular bounding-box) by adding more weight to the corresponding pixels but were not able to get any significant improvements. Our earlier experiments with adversarial loss \cite{Goodfellow14} were also inconclusive. As the residual flow predicted by the $ 1\times $-scale module tends to be quite noisy, we attempted to smoothen the flow-field by imposing a total variation penalty. Unfortunately, this resulted in a slightly worse $ \ell_2 $-loss on the test set.

\subsubsection{Sampling training pairs.}
We found that biasing the selection process for more difficult and unusual head poses and bigger redirection angles improved the results.  For this reason, we used the following sampling scheme aimed at reducing the dataset imbalance. We split all possible correction angles (that is, the range between $ -30\degree $ and $ 30\degree $) into $ 15 $ bins. A set of samples falling into a bin is further divided into ``easy'' and ``hard'' subsets depending on the input's \emph{tilt} angle (an angle between the segment connecting two most distant eye feature points and the horizontal baseline). A sample is considered to be ``hard'' if its tilt is $ \geqslant 8\degree $. This subdivision helps to identify training pairs corresponding to the rare head poses.
We form a training batch by picking 4 correction angle bins uniformly at random and sampling 24 ``easy'' and 8 ``hard'' examples for each of the chosen bins.

\subsection{Quantitative evaluation}
We evaluate our approach on our dataset. We randomly split the initial set of subjects into a development (26 persons) and a test (7 persons) sets. Several methods were compared using the mean square error (MSE) between the synthesized and the ground-truth images extracted using the procedure described in \sect{methods:localization}.

\subsubsection{Models.}
We consider 6 different models:
\begin{enumerate}
  \item A system based on Structured Random Forests (\emph{RF}) proposed in \cite{Kononenko15}. We train it for $ 15\degree $ redirection only using the reference implementation.
  \item A single-scale (\emph{SS} ($ 15\degree $ only)) version of our method with a single warping module operating on the original image scale that is trained for $ 15\degree $ redirection only.
  \item A single-scale (\emph{SS}) version of our method with a single warping module operating on the original image scale.
  \item A multi-scale (\emph{MS}) network without coarse warping. It processes inputs on two scales and uses features from both scales to predict the final warping transformation.
  \item A coarse-to-fine warping-based system described in \sect{model} (\emph{CFW}).
  \item A coarse-to-fine warping-based system with a lightness correction module (\emph{CFW + LCM}).
\end{enumerate}
The latter four models are trained for the task of vertical gaze redirection in the range. We call such models \emph{unified} (as opposed to single angle correction systems). 

\subsubsection{$ 15\degree $ correction.}
\begin{figure}
  \centering
  \setlength\figureheight{7.5cm}
  \setlength\figurewidth{12cm}
  \input{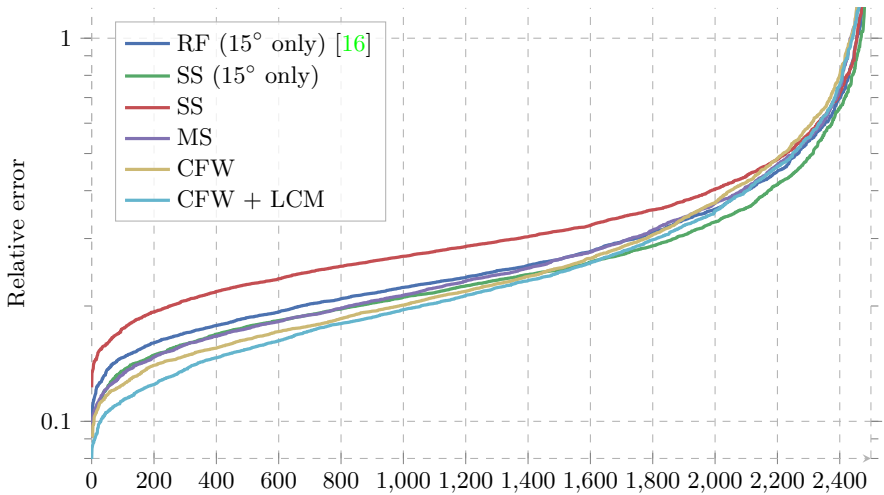}
  \caption{Ordered errors for $ 15\degree $ redirection. Our multi-scale models (MS, CFW, CFW + LCM) show results that are comparable or superior the Random Forests (RF) \cite{Kononenko15}.}
  \label{fig:15_error_curve}
\end{figure}

In order to have the common ground with the existing systems, we first restrict ourselves to the case of $ 15\degree $ gaze correction. Following \cite{Kononenko15}, we present a graph of sorted normalized errors (\fig{15_error_curve}), where all errors are divided by the MSE obtained by an input image and then the errors on the test set are sorted for each model.

It can be seen that the unified multi-scale models are, in general, comparable or superior to the RF-based approach in \cite{Kononenko15}. Interestingly, the lightness adjustment extension (\sect{methods:palette}) is able to show quite significant improvements for the samples with low MSE. Those are are mostly cases similar to shown in \fig{palette_examples}. It is also worth noting that the single-scale model trained for this specific correction angle consistently outperforms \cite{Kononenko15}, demonstrating the power of the proposed architecture. However, we note that results of the methods can be improved using additional registration procedure, one example of which is described in \sect{experiments:registration}.

\subsubsection{Arbitrary vertical redirection.}
\begin{figure}
  \centering
  \setlength\figureheight{7.5cm}
  \setlength\figurewidth{12cm}
  \input{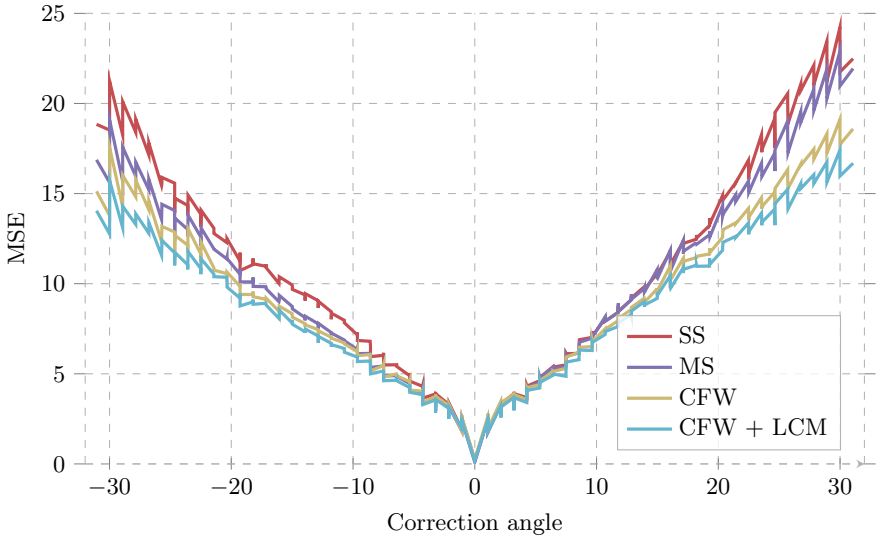}
  \caption{Distribution of errors over different correction angles.}
  \label{fig:error_dist}
\end{figure}

We also compare different variants of unified networks and plot the error distribution over different redirection angles (\fig{error_dist}). For small angles, all the methods demonstrate roughly the same performance, but as we increase the amount of correction, the task becomes much harder (which is reflected by the growing error) revealing the difference between the models. Again, the best results are achieved by the palette model, which is followed by the multi-scale networks making use of coarse warping.

\subsection{Perceptual quality}

\newlength{\diewidth}
\setlength{\diewidth}{60mm}
\begin{figure*}
\centering
\newcolumntype{C}[1]{>{\centering\let\newline\\\arraybackslash\hspace{0pt}}m{#1}}
\begin{tabular}{C{11mm}C{11mm}C{11mm}C{11mm}C{11mm}C{11mm}C{11mm}C{11mm}C{11mm}C{11mm}}
Input & RF & CFW & +LCM & GT & Input & RF & CFW & +LCM & GT \\
\multicolumn{5}{l}{\includegraphics[width=\diewidth]{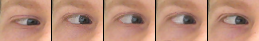}}&
\multicolumn{5}{l}{\includegraphics[width=\diewidth]{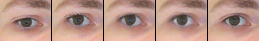}}\\

\multicolumn{5}{l}{\includegraphics[width=\diewidth]{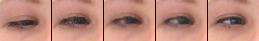}}&
\multicolumn{5}{l}{\includegraphics[width=\diewidth]{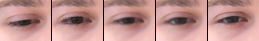}}\\

\multicolumn{5}{l}{\includegraphics[width=\diewidth]{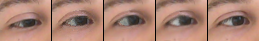}}&
\multicolumn{5}{l}{\includegraphics[width=\diewidth]{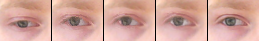}}\\

\multicolumn{5}{l}{\includegraphics[width=\diewidth]{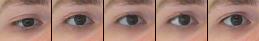}}&
\multicolumn{5}{l}{\includegraphics[width=\diewidth]{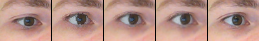}}\\

\multicolumn{5}{l}{\includegraphics[width=\diewidth]{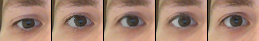}}&
\multicolumn{5}{l}{\includegraphics[width=\diewidth]{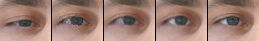}}\\

\multicolumn{5}{l}{\includegraphics[width=\diewidth]{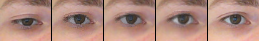}}&
\multicolumn{5}{l}{\includegraphics[width=\diewidth]{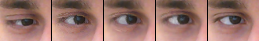}}\\

\end{tabular}

\caption{Sample results on a hold-out. The full version of our model (CFW+LCM) outperforms other methods.}
\label{fig:dies}
\end{figure*}

We demonstrate the results of redirection on $15$ degrees upwards in the \fig{dies}. CFW-based systems produce the results visually closer to the ground truth, than RF. The effect of the lightness correction is pronounced: on the input image with the lack of white Random Forest and CFW fail to get output with sufficient eye-white and copy-paste red pixels instead, whereas CFW+LCM achieve good correspondence with the ground-truth. However, the downside effect of the LCM could be blurring/lower contrast because of the multiplication procedure \eq{LCM}.

\begin{table}
\centering
\caption{\textbf{User assessment for the photorealism of the results for the four methods.} During the session, each of the 16 test subjects observed 40 instances of results of each method embedded within 3 real eye images. The participants were asked to click on the resynthesized image in as little time as they could. The first three parts of the table specify the number of correct guesses (the smaller the better). The last line indicates the mean time needed to make a guess (the larger the better). Our full system (coarse-to-fine warping and lightness correction) dominated the performance.}
\label{tab:user}
\begin{tabular}{|l|c|c|c|c|}
\hline
\cellcolor[HTML]{EFEFEF}               & \multicolumn{1}{l|}{\cellcolor[HTML]{EFEFEF}\textbf{Random Forest}} & \multicolumn{1}{l|}{\cellcolor[HTML]{EFEFEF}\textbf{Single Scale}} & \multicolumn{1}{l|}{\cellcolor[HTML]{EFEFEF}\textbf{CFW}} & \multicolumn{1}{l|}{\cellcolor[HTML]{EFEFEF}\textbf{CFW+LCM}} \\ \hline
\multicolumn{5}{|c|}{\cellcolor[HTML]{FDE7E5}Correctly guessed (out of 40)}                                                                                                                                                                                                                                   \\ \hline
\cellcolor[HTML]{FDE7E5}Mean           & \cellcolor[HTML]{FDE7E5}36.1                                        & \cellcolor[HTML]{FDE7E5}33.8                                       & \cellcolor[HTML]{FDE7E5}28.8                              & \cellcolor[HTML]{FDE7E5}\textbf{25.3}                         \\ \hline
\cellcolor[HTML]{FDE7E5}Median         & \cellcolor[HTML]{FDE7E5}37                                          & \cellcolor[HTML]{FDE7E5}35                                         & \cellcolor[HTML]{FDE7E5}29                                & \cellcolor[HTML]{FDE7E5}\textbf{25}                           \\ \hline
\cellcolor[HTML]{FDE7E5}Max            & \cellcolor[HTML]{FDE7E5}40                                          & \cellcolor[HTML]{FDE7E5}39                                         & \cellcolor[HTML]{FDE7E5}38                                & \cellcolor[HTML]{FDE7E5}\textbf{34}                           \\ \hline
\cellcolor[HTML]{FDE7E5}Min            & \cellcolor[HTML]{FDE7E5}26                                          & \cellcolor[HTML]{FDE7E5}22                                         & \cellcolor[HTML]{FDE7E5}20                                & \cellcolor[HTML]{FDE7E5}\textbf{16}                           \\ \hline
\multicolumn{5}{|c|}{\cellcolor[HTML]{DEFDDE}Correctly guessed within 2 seconds (out of 40)}                                                                                                                                                                                                                  \\ \hline
\cellcolor[HTML]{DEFDDE}Mean           & \cellcolor[HTML]{DEFDDE}26.4                                        & \cellcolor[HTML]{DEFDDE}21.1                                       & \cellcolor[HTML]{DEFDDE}11.7                              & \cellcolor[HTML]{DEFDDE}\textbf{8.0}                          \\ \hline
\cellcolor[HTML]{DEFDDE}Median         & \cellcolor[HTML]{DEFDDE}28.5                                        & \cellcolor[HTML]{DEFDDE}20.5                                       & \cellcolor[HTML]{DEFDDE}10                                & \cellcolor[HTML]{DEFDDE}\textbf{8}                            \\ \hline
\cellcolor[HTML]{DEFDDE}Max            & \cellcolor[HTML]{DEFDDE}35                                          & \cellcolor[HTML]{DEFDDE}33                                         & \cellcolor[HTML]{DEFDDE}23                                & \cellcolor[HTML]{DEFDDE}\textbf{17}                           \\ \hline
\cellcolor[HTML]{DEFDDE}Min            & \cellcolor[HTML]{DEFDDE}13                                          & \cellcolor[HTML]{DEFDDE}11                                         & \cellcolor[HTML]{DEFDDE}3                                 & \cellcolor[HTML]{DEFDDE}\textbf{0}                            \\ \hline
\multicolumn{5}{|c|}{\cellcolor[HTML]{ECF4FF}Correctly guessed within 1 second (out of 40)}                                                                                                                                                                                                                   \\ \hline
\cellcolor[HTML]{ECF4FF}Mean           & \cellcolor[HTML]{ECF4FF}8.1                                         & \cellcolor[HTML]{ECF4FF}4.4                                        & \cellcolor[HTML]{ECF4FF}1.6                               & \cellcolor[HTML]{ECF4FF}\textbf{1.1}                          \\ \hline
\cellcolor[HTML]{ECF4FF}Median         & \cellcolor[HTML]{ECF4FF}6                                           & \cellcolor[HTML]{ECF4FF}3                                          & \cellcolor[HTML]{ECF4FF}\textbf{1}                        & \cellcolor[HTML]{ECF4FF}\textbf{1}                            \\ \hline
\cellcolor[HTML]{ECF4FF}Max            & \cellcolor[HTML]{ECF4FF}20                                          & \cellcolor[HTML]{ECF4FF}15                                         & \cellcolor[HTML]{ECF4FF}7                                 & \cellcolor[HTML]{ECF4FF}\textbf{5}                            \\ \hline
\cellcolor[HTML]{ECF4FF}Min            & \cellcolor[HTML]{ECF4FF}\textbf{0}                                  & \cellcolor[HTML]{ECF4FF}\textbf{0}                                 & \cellcolor[HTML]{ECF4FF}\textbf{0}                        & \cellcolor[HTML]{ECF4FF}\textbf{0}                            \\ \hline
\multicolumn{5}{|c|}{\cellcolor[HTML]{FBFBE6}Mean time to make a guess}                                                                                                                                                                                                                                       \\ \hline
\cellcolor[HTML]{FBFBE6}Mean time, sec & \cellcolor[HTML]{FBFBE6}1.89                                        & \cellcolor[HTML]{FBFBE6}2.30                                       & \cellcolor[HTML]{FBFBE6}3.60                              & \cellcolor[HTML]{FBFBE6}\textbf{3.96}                         \\ \hline
\end{tabular}
\end{table}

\subsubsection{User study} To confirm the improvement corresponding to different aspects of the proposed models, which may not be adequately reflected by $ \ell_2 $-measure, we performed an informal user study enrolling $16$ subjects unrelated to computer vision and comparing four methods (RF, SS, CFW, CFW+LCM). 
Each user was shown $160$ quadruplets of images, and in each quadruplet one of the images was obtained by re-synthesis with one of the methods, while the remaining three were unprocessed real images of eyes. $40$ randomly sampled results from each of the compared methods were thus embedded. When a quadruplet was shown, the task of the subject was to click on the artificial (re-synthesized) image as quickly as possible. For each method, we then recorded the number of correct guesses out of $40$ (for an ideal method the expected number would be $10$, and for a very poor one it would be $40$). We also recorded the time that the subject took to decide on each quadruplet (better method would take a longer time for spotting). Table~\ref{tab:user} shows results of the experiment. Notably, here the gap between methods is much wider then it might seem from the MSE-based comparisons, with CFW+LCM method outperforming others very considerably, especially when taking into account the timings.

\subsubsection{Horizontal redirection.} While most of our experiments were about vertical gaze redirection, the same models can be trained to redirect the gaze horizontally (and, with trivial generalization, by a 2D family of angles). In \fig{horizontal}, we provide qualitative results of CFW+LCM for horizontal redirection. Some examples showing the limitations of our method are given. The limitations are concerned with cases with severe disocclusions, where large areas have to be filled by the network.

We provide more qualitative results on the project webpage \cite{Website}.

\subsection{Incorporating registration}
\label{sect:experiments:registration}

We found that results can be further perceptually improved (see \cite{Website}) if the objective is slightly modified to take into account misalignment between inputs and ground-truth images. To that end, we enlarge the bounding-box $ \mathcal{B} $ that we use to extract the output image of a training pair by $ k = 3 $ pixels in all the directions. Given that now $ O_{gt} $ has the size of $ (H + 2k) \times (W + 2k) $, the new objective is defined as:
\begin{equation}
  \mathcal{L}(O_{\text{output}}, O_{\text{gt}}) = \min_{i, j} \mbox{dist} \left(O_{\text{output}}, O_{\text{gt}}[i : i + H, j : j + W] \right) \, ,
\end{equation}
where $ \text{dist}(\cdot) $ can be either $ \ell_2 $ or $ \ell_1 $-distance (the latter giving slightly sharper results), and $ O_{\text{gt}}[i : i + H, j : j + W] $ corresponds to a $ H \times W $ crop of $ O_{\text{gt}} $ with top left corner at the position $ (i, j) $. Being an alternative to the offline registration of input/ground-truth pairs \cite{Kononenko15} which is computationally prohibitive in large-scale scenarios, this small trick greatly increases robustness of the training procedure against small misalignments in a training set. 

\newcommand{\horrow}[1]{%
\begin{tikzpicture}
\foreach \x[evaluate=\x as \emx using int(4 - \x)] in {0,...,4} {
  \node [anchor=north west,inner sep=0] at (\x * 2.2cm, 0) {\pgfimage[width=2.2cm]{./figures/images/horizontal/#1_00\x.png}};
}
\end{tikzpicture}%
}
\tikzexternaldisable
\begin{figure*}
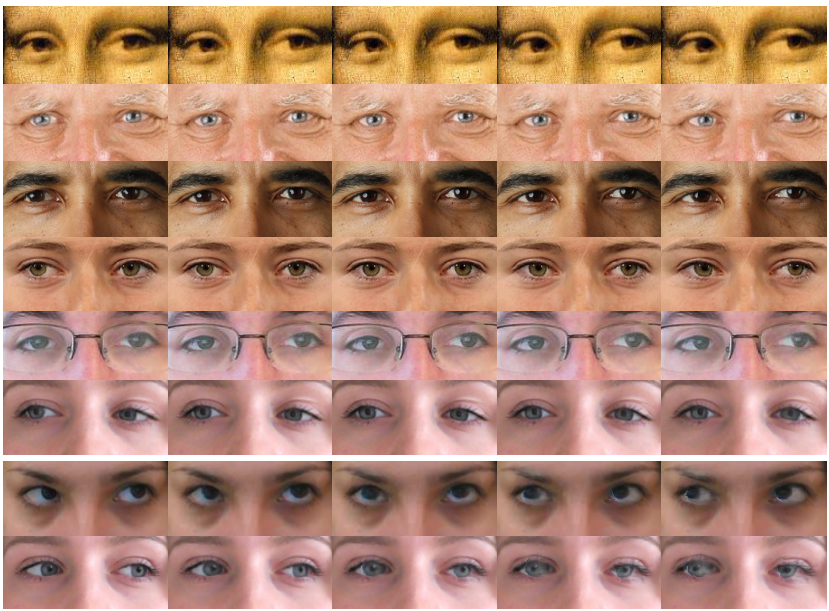

\centering
\setlength{\tabcolsep}{0pt}
\renewcommand{\arraystretch}{0}
\begin{tabular}{ccccc}
\horrow{joconde_ms3_fuse_palette_xy} \\
\horrow{harold_ms3_fuse_palette_xy} \\
\horrow{obama_ms3_fuse_palette_xy} \\
\horrow{sharapova_ms3_fuse_palette_xy} \\
\horrow{dima_ms3_fuse_palette_xy} \\
\horrow{test1_ms3_fuse_palette_xy} \\
\noalign{\vskip 1mm} 
\horrow{diana2_ms3_fuse_palette_xy} \\
\horrow{extreme_test1_ms3_fuse_palette_xy}
\end{tabular}
\caption{{\bf Horizontal redirection} with a model trained for both vertical and horizontal gaze redirection. For the first six rows the angle varies from $ -15\degree$ to $ 15\degree$ relative to the central (input) image. The last two rows push the redirection to extreme angles (up to $45\degree$) breaking our model down.}
\label{fig:horizontal}
\end{figure*}
\tikzexternalenable

\section{Discussion}

We have suggested a method for realistic gaze redirection, allowing to change gaze continuously in a certain range. At the core of our approach is the prediction of the warping field using a deep convolutional network. We embed redirection angle and feature points as image-sized maps and suggest ``fully-convolutional'' coarse-to-fine architecture of warping modules. In addition to warping, photorealism is increased using lightness correction module. Quantitative comparison of MSE-error, qualitative examples and a user study show the advantage of suggested techniques and the benefit of their combination within an end-to-end learnable framework.

Our system is reasonably robust against different head poses (e.g., see \fig{palette_examples}) and deals correctly with the situations where a person wears glasses (see \cite{Website}). Most of the failure modes (e.g., corresponding to extremely tilted head poses or large redirection angles involving disocclusion of the different parts of an eye) are not inherent to the model design and can be addressed by augmenting the training data with appropriate examples.

We concentrated on gaze redirection, although our approach might be extended for other similar tasks, e.g.\ re-synthesis of faces. In contrast with\\autoencoders-based approach, our architecture does not compress data to a representation with lower explicit or implicit dimension, but directly transforms the input image. Our method thus might be better suited for fine detail preservation, and less prone to the ``regression-to-mean'' effect.

The computational performance of our method is up to $20$ fps on a mid-range consumer GPU (NVIDIA GeForce-750M), which is however slower than the competing method of \cite{Kononenko15}, which is able to achieve similar speed on CPU. Our models are however much more compact than forests from \cite{Kononenko15} ($250$ Kb vs $30$-$60$ Mb in our comparisons), while also being universal. We are currently working on the unification of the two approaches.

Speed optimization of the proposed system is another topic for future work. Finally, we plan to further investigate non-standard loss functions for our architectures (e.g. the one proposed in \sect{experiments:registration}), as the $ \ell_2 $-loss is not closely enough related to perceptual quality of results (as highlighted by our user study).

\section*{Acknowledgements}
We would like to thank Leonid Ekimov for sharing the results of his work on applying auto-encoders for gaze correction. We are also grateful to all the Skoltech students and employees who agreed to participate in the dataset collection and in the user study. This research is supported by the Skoltech Translational Research and Innovation Program.

\FloatBarrier
\newpage

\bibliographystyle{splncs03}
\bibliography{references}

\begin{appendices}
\renewcommand{\thesection}{\appendixname~\Alph{section}}%
\section{Drawbacks of conventional architectures}

\begin{figure}[t!]
    \centering
    \begin{tabular}{ccc}
    \includegraphics[width=3.3cm]{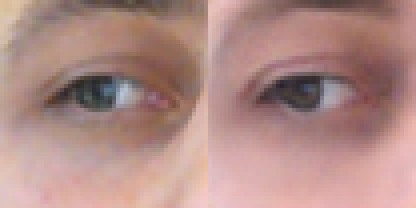}&
    \includegraphics[width=3.3cm]{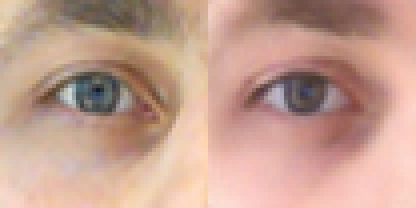}&
    \includegraphics[width=3.3cm]{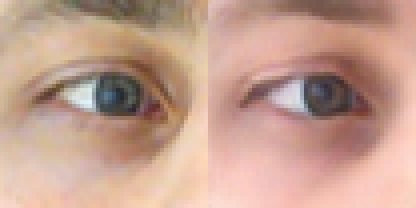}
    \end{tabular}
    \caption{Examples of reconstructions produced by a modern \textbf{encoder-decoder} architecture (following the approach in \cite{Kulkarni15,Reed15,Ghodrati15}) trained on our data. In each pair, the left image is the input and the right is the output. Despite our efforts, a noticeable loss of fine-scale details and ``regression-to-mean'' effect make the result not good enough for most applications of gaze manipulation. Similar problems can be observed in \cite{Kulkarni15,Reed15,Ghodrati15}.}
    \label{fig:autoenc}
\end{figure}

In order to determine the applicability of conventional generative architectures for gaze correction, we used our data to train several auto-encoders. The best model has $ 200 $-dimensional latent space and consists of several convolutional and fully-connected layers in the encoder and the decoder. We use a combination of $ \ell_2 $ and GAN~\cite{Goodfellow14} losses to achieve the best possible results. Unfortunately, due to inherently lossy encoding procedure, the model exhibits noticeable fine-scale details dropping and ``regression-to-mean'' effect (see \fig{autoenc}). That makes incorporation of such kind of approach into a gaze correction system problematic. 

\section{Details of the proposed method}

Here we give a more detailed view of the architecture that we use in our gaze correction system.

\subsection{Warping stage}

A flowchart of the pipeline {\it without} the lightness correction module is depicted in \fig{ms3_arch}. To allow for more interaction between the input data and the correction angle embedding, we choose not to perform late fusion \cite{Ghodrati15} and feed the embedding vector as an additional input to the warping network. More concretely, we replicate the $ 16 $-dimensional $ \Fsub{angle} (\alpha) $ for every spatial location creating a tensor of size $ 16 \times H \times W $  which is then concatenated to the rest of the input ({\it pink} triangle in \fig{ms3_arch:high_level}). The architectures of the two warping modules are same (modulo the number of input maps) and are shown in \fig{ms3_arch:conv_layers}.

\subsection{Lightness correction module}
\begin{figure*}
\centering
\subfigure[Architecture.\label{fig:palette:arch}]{
  \scalebox{0.33}{\begin{tikzpicture}[
  black!50,
  text=black!80,
  fnode/.style={
    align=center,
    rectangle,minimum height=40pt,minimum width=80pt,rounded corners=2pt,
    inner sep=5pt,
    line width=3pt,
    fill=pnodefill,draw=pnodedraw,
    font=\ttfamily\Large},
  dnode/.style={
    fnode,trapezium,trapezium left angle=70, trapezium right angle=110,trapezium stretches},
  inode/.style={
    dnode,fill=inodefill,draw=inodedraw,minimum width=110pt},
  pnode/.style={
    fnode,fill=pnodefill,draw=pnodedraw},
  onode/.style={
    dnode,fill=onodefill,draw=onodedraw},
  mnode/.style={
    pnode,isosceles triangle,isosceles triangle stretches,minimum width=35pt,minimum height=35pt},
  smnode/.style={
    mnode,fill=mnodefill,draw=mnodedraw},
  flow/.style={
    -latex,shorten >=1pt,line width=3pt,line cap=round,rounded corners=2pt,draw=pnodedraw,draw=#1},
  branch/.style={
    circle,inner sep=0pt,minimum size=8pt,fill=pnodedraw,draw=pnodedraw,fill=#1,draw=#1},
  adder/.style={
    circle,minimum size=35pt,line width=3pt,fill=pnodefill,draw=pnodedraw},
  vhedge/.style={
    rounded corners,to path=|- (\tikztotarget)}]
  \newcommand{\mytrap}[5]{%
    \node (#1)  [anchor=center,#2] {\phantom{#4}\\\phantom{#5}}; \node[anchor=center,font=\ttfamily\Large] at (#1.center) {#3};
  }
  \matrix[row sep=20pt,column sep=19pt,ampersand replacement=\&] {
    \node (05x_feats)     [inode]  {$ 0.5\times $-scale\\features}; \&
    \node (dummy_top)              {}; \\

  	\node (1x_feats)      [inode]  {$ 1\times $-scale\\features}; \&
  	\& 
  	\node (merge1)        [mnode]  {}; \&
	\node (conv_layers)   [pnode]  {conv\\layers}; \& 
	\node (weights)       [dnode]  {per-pixel\\weights}; \&
	\node (mulred)        [adder]  {\Huge $ \bullet $}; \&
	\node (cor_out)       [onode]  {corrected\\out}; \\
	
    \mytrap{out}{inode}{out}{$ 0.5\times $-scale}{features} \&
	\&
	\node (merge2)        [smnode] {}; \\
	
    \mytrap{palette}{inode}{palette}{$ 0.5\times $-scale}{features} \&
	\node (dummy_bottom)           {}; \\
  };
  
  \begin{scope}[on background layer]
    \draw [flow=inodedraw] (05x_feats) -- (dummy_top.center) |- ($ (merge1.west) + (0,10pt) $);
    
    \path (1x_feats) edge[flow=inodedraw] (merge1);
    \path (merge1) edge[flow] (conv_layers);
    \path (conv_layers) edge[flow] (weights);
    \path (weights) edge[flow] (mulred);
    \path (mulred) edge[flow] (cor_out);
    
    \path (out) edge[flow=inodedraw] (merge2);
    \draw [flow=mnodedraw] (merge2) -| (mulred);
    
    \draw [flow=inodedraw] (palette) -- (dummy_bottom.center) |- ($ (merge2.west) - (0,10pt) $);
  \end{scope}
\end{tikzpicture}}}
\subfigure[Convolutional layers.\label{fig:palette:conv_layers}]{
  \scalebox{0.33}{\begin{tikzpicture}[
  black!50,
  text=black!80,
  fnode/.style={
    align=center,
    rectangle,minimum height=40pt,minimum width=40pt,rounded corners=2pt,
    inner sep=5pt,
    line width=3pt,
    fill=pnodefill,draw=pnodedraw,
    font=\ttfamily\Large},
  dnode/.style={
    fnode,trapezium,trapezium left angle=70, trapezium right angle=110,trapezium stretches},
  inode/.style={
    dnode,fill=inodefill,draw=inodedraw},
  pnode/.style={
    fnode,fill=pnodefill,draw=pnodedraw},
  pnode2/.style={
    fnode,fill=pnodefill!50!white,draw=pnodedraw!50!white},
  onode/.style={
    dnode,fill=onodefill,draw=onodedraw},
  smnode/.style={
    mnode,fill=mnodefill,draw=mnodedraw},
  flow/.style={
    -latex,shorten >=1pt,line width=3pt,line cap=round,rounded corners=2pt,draw=pnodedraw,draw=#1},
  flow2/.style={
    flow,draw=pnodedraw!50!white}]
  \matrix[row sep=20pt,column sep=19pt,ampersand replacement=\&] {
	\node (conv1) [pnode]  {conv 1x1\\8 maps}; \& 
	\node (bn1)   [pnode]  {batch\\norm}; \&
	\node (relu1) [pnode]  {ReLU}; \&
	
	\node (conv2) [pnode2] {conv 1x1\\8 maps}; \& 
	\node (bn2)   [pnode2] {batch\\norm}; \&
	\node (relu2) [pnode2] {ReLU}; \&
	
	\node (conv3) [pnode]  {conv 3x3\\32 maps}; \& 
	\node (sm)    [pnode]  {spatial\\softmax}; \\
  };
  
  \begin{scope}[on background layer]
    \path (conv1) edge[flow] (bn1);
    \path (bn1) edge[flow] (relu1);
    \path (relu1) edge[flow] (conv2);
    
    \path (conv2) edge[flow2] (bn2);
    \path (bn2) edge[flow2] (relu2);
    \path (relu2) edge[flow2] (conv3);
    
    \path (conv3) edge[flow] (sm);
  \end{scope}
\end{tikzpicture}}}
\caption{\textbf{Lightness Correction Module} increases lightness of selected regions. \ref{fig:palette:arch} shows the actual architecture of the module. Multi-scale features are processed by the convolutional neural network presented in \ref{fig:palette:conv_layers}.}
\label{fig:palette}
\end{figure*}
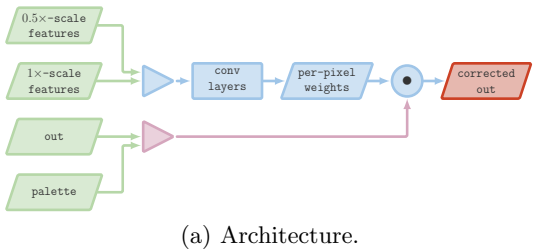
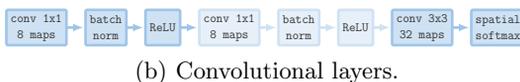

\fig{palette} shows the actual architecture for the lightness correction module. Per-pixel weights are predicted based on the internal activations (the third convolutional layer) of the warping modules ($ 0.5 \times $-\texttt{scale} and $ 1 \times $-\texttt{scale features} in the scheme respectively).

\begin{figure*}
\centering
\subfigure[High-level pipeline\label{fig:ms3_arch:high_level}]{
  \scalebox{0.33}{\begin{tikzpicture}[
  black!50,
  text=black!80,
  fnode/.style={
    align=center,
    rectangle,minimum height=40pt,minimum width=80pt,rounded corners=2pt,
    inner sep=5pt,
    line width=3pt,
    fill=pnodefill,draw=pnodedraw,
    font=\ttfamily\Large},
  dnode/.style={
    fnode,trapezium,trapezium left angle=70, trapezium right angle=110,trapezium stretches},
  inode/.style={
    dnode,fill=inodefill,draw=inodedraw},
  pnode/.style={
    fnode,fill=pnodefill,draw=pnodedraw},
  onode/.style={
    dnode,fill=onodefill,draw=onodedraw},
  mnode/.style={
    pnode,isosceles triangle,isosceles triangle stretches,minimum width=35pt,minimum height=35pt},
  smnode/.style={
    mnode,fill=mnodefill,draw=mnodedraw},
  flow/.style={
    -latex,shorten >=1pt,line width=3pt,line cap=round,rounded corners=2pt,draw=pnodedraw,draw=#1},
  branch/.style={
    circle,inner sep=0pt,minimum size=8pt,fill=pnodedraw,draw=pnodedraw,fill=#1,draw=#1},
  vhedge/.style={
    rounded corners,to path=|- (\tikztotarget)}]
  \newcommand{\mytrap}[5]{%
    \node (#1)  [anchor=center,#2] {\phantom{#4}\\\phantom{#5}}; \node[anchor=center,font=\ttfamily\Large] at (#1.center) {#3};
  }
  \matrix[row sep=5pt,column sep=19pt,ampersand replacement=\&] {
  	\mytrap{angle}{inode}{angle}{16}{units} \&
	\node (embed_angle)     [pnode]                   {embed\\angle}; \& 
	\node (angle_feat)      [dnode]                   {16\\units}; \&
	\node (premerge1_1)                               {}; \&
	\& \& 
	\node (process_05scale) [pnode]                   {process\\$ 0.5\times $ scale}; \&
	\node (coarse_out)      [dnode]                   {coarse\\out}; \&
	\node (premerge2_1)                               {}; \\
	
	\& \& \& \&
	\node (merge1)          [smnode]                  {}; \&
	\node (postmerge1)      [branch=mnodedraw]		  {}; \&
	\& \& \&
	\node (merge2)          [mnode]                   {}; \&
	\node (process_1scale)  [pnode]                   {process\\$ 1\times $ scale}; \&
	\mytrap{out}{onode}{out}{16}{units} \\
  
    \& \&
    \mytrap{input}{inode}{input}{16}{units} \&
    \node (premerge1_2)																{}; \&
    \& \& \&
	\node (coarse_flow)     [dnode]                   {coarse\\flow}; \\
  };
  
  \begin{scope}[on background layer]
    \path (angle) edge[flow=inodedraw] (embed_angle);
    \path (embed_angle) edge[flow] (angle_feat);
    
    \draw [flow] (angle_feat.east) -- (premerge1_1.center) |- ($ (merge1.west) + (0,10pt) $);
    \draw [flow=inodedraw] (input.east) -- (premerge1_2.center) |- ($ (merge1.west) - (0,10pt) $);
    
    \path (postmerge1) edge[flow=mnodedraw,vhedge] (process_05scale);
    \path (process_05scale) edge[flow] (coarse_out);
    \path (process_05scale) edge[flow,vhedge] (coarse_flow);
    
    \draw [flow] (coarse_out.east) -- (premerge2_1.center) |- ($ (merge2.west) + (0,10pt) $);
    \draw [flow] (coarse_flow.east) -| (process_1scale.south);
    \path (merge1) edge[flow=mnodedraw] (postmerge1);
    \draw[flow=mnodedraw] (postmerge1) |- ($ (merge2.west) - (0,10pt) $);
    \path (merge2) edge[flow] (process_1scale);
    \path (process_1scale) edge[flow] (out);
  \end{scope}
\end{tikzpicture}}}
\subfigure[$ 0.5\times $-scale processing module\label{fig:ms3_arch:05_scale}]{
  \scalebox{0.33}{\begin{tikzpicture}[
  black!50,
  text=black!80,
  fnode/.style={
    align=center,
    rectangle,minimum height=40pt,minimum width=80pt,rounded corners=2pt,
    inner sep=5pt,
    line width=3pt,
    fill=pnodefill,draw=pnodedraw,
    font=\ttfamily\Large},
  dnode/.style={
    fnode,trapezium,trapezium left angle=70, trapezium right angle=110,trapezium stretches},
  inode/.style={
    dnode,fill=inodefill,draw=inodedraw},
  pnode/.style={
    fnode,fill=pnodefill,draw=pnodedraw},
  onode/.style={
    dnode,fill=onodefill,draw=onodedraw},
  mnode/.style={
    pnode,isosceles triangle,isosceles triangle stretches,minimum width=35pt,minimum height=35pt},
  smnode/.style={
    mnode,fill=mnodefill,draw=mnodedraw},
  flow/.style={
    -latex,shorten >=1pt,line width=3pt,line cap=round,rounded corners=2pt,draw=pnodedraw,draw=#1},
  branch/.style={
    circle,inner sep=0pt,minimum size=8pt,fill=pnodedraw,draw=pnodedraw,fill=#1,draw=#1},
  vhedge/.style={
    rounded corners,to path=|- (\tikztotarget)}]
  \newcommand{\mytrap}[5]{%
    \node (#1)  [anchor=center,#2] {\phantom{#4}\\\phantom{#5}}; \node[anchor=center,font=\ttfamily\Large] at (#1.center) {#3};
  }
  \matrix[row sep=20pt,column sep=19pt,ampersand replacement=\&] {
    \&
    \node (dummy_top)                             {}; \\
  
  	\node (input_maps)  [inode]                   {input\\maps}; \&
  	\node (postinput)   [branch=inodedraw]        {}; \&
  	\node (downsample)  [pnode]                   {downsample}; \&
	\node (conv_layers) [pnode]                   {conv\\layers}; \& 
	\node (upsample)    [pnode]                   {upsample}; \&
	\mytrap{flow}{onode}{flow}{16}{units} \&
	\node (bilinear)    [pnode]					  {bilinear\\sampler}; \&
	\mytrap{out}{onode}{out}{16}{units} \&
	\node (dummy_out)                             {}; \\
	
	\& \& \& \& \& \& \& \&
	\node (dummy_flow)                            {}; \\
  };
  
  \begin{scope}[on background layer]
    \path (input_maps) edge[flow=inodedraw] (downsample);
    \path (downsample) edge[flow] (conv_layers);
    \draw [flow=inodedraw] (postinput) -- (dummy_top.center) -| (bilinear.north);
    
    \path (conv_layers) edge[flow] (upsample);
    \path (upsample) edge[flow] (flow);
    \path (flow) edge[flow=onodedraw] (bilinear);
    \path (bilinear) edge[flow] (out);
    \path (out) edge[flow=onodedraw] (dummy_out);
    
    \draw [flow=onodedraw] (flow.south) |- (dummy_flow.center);
  \end{scope}
\end{tikzpicture}}}
\subfigure[$ 1\times $-scale processing module\label{fig:ms3_arch:1_scale}]{
  \scalebox{0.33}{\begin{tikzpicture}[
  black!50,
  text=black!80,
  fnode/.style={
    align=center,
    rectangle,minimum height=40pt,minimum width=80pt,rounded corners=2pt,
    inner sep=5pt,
    line width=3pt,
    fill=pnodefill,draw=pnodedraw,
    font=\ttfamily\Large},
  dnode/.style={
    fnode,trapezium,trapezium left angle=70, trapezium right angle=110,trapezium stretches},
  inode/.style={
    dnode,fill=inodefill,draw=inodedraw},
  pnode/.style={
    fnode,fill=pnodefill,draw=pnodedraw},
  onode/.style={
    dnode,fill=onodefill,draw=onodedraw},
  mnode/.style={
    pnode,isosceles triangle,isosceles triangle stretches,minimum width=35pt,minimum height=35pt},
  smnode/.style={
    mnode,fill=mnodefill,draw=mnodedraw},
  flow/.style={
    -latex,shorten >=1pt,line width=3pt,line cap=round,rounded corners=2pt,draw=pnodedraw,draw=#1},
  branch/.style={
    circle,inner sep=0pt,minimum size=8pt,fill=pnodedraw,draw=pnodedraw,fill=#1,draw=#1},
  adder/.style={
    circle,minimum size=35pt,line width=3pt,fill=pnodefill,draw=pnodedraw},
  vhedge/.style={
    rounded corners,to path=|- (\tikztotarget)}]
  \newcommand{\mytrap}[5]{%
    \node (#1)  [anchor=center,#2] {\phantom{#4}\\\phantom{#5}}; \node[anchor=center,font=\ttfamily\Large] at (#1.center) {#3};
  }
  \matrix[row sep=20pt,column sep=19pt,ampersand replacement=\&] {
    \&
    \node (dummy_top)                             {}; \\
  
  	\node (input_maps)  [inode]                   {input\\maps}; \&
  	\node (postinput)   [branch=inodedraw]        {}; \&
	\node (conv_layers) [pnode]                   {conv\\layers}; \& 
	\node (flow)        [dnode]                   {res\\flow}; \&
	\node (bilinear)    [pnode]					  {bilinear\\sampler}; \&
	\mytrap{out}{onode}{out}{16}{units} \&
	\node (dummy_out)                             {}; \\
	
	\node (coarse_flow) [inode]                   {coarse\\flow}; \&
	\&
	\node (postcoarse)  [branch=inodedraw]        {}; \&
	\node (add_res)     [adder]                   {\textbf{\Huge+}}; \\
  };
  
  \begin{scope}[on background layer]
    \path (input_maps) edge[flow=inodedraw] (conv_layers);
    \draw [flow=inodedraw] (postinput) -- (dummy_top.center) -| (bilinear.north);
    
    \path (conv_layers) edge[flow] (flow);
    \path (bilinear) edge[flow] (out);
    \path (out) edge[flow=onodedraw] (dummy_out);
    \path (flow) edge[flow] (add_res);
    
    \path (coarse_flow) edge[flow=inodedraw] (add_res);
    \path (postcoarse) edge[flow=inodedraw] (conv_layers);
    \draw [flow] (add_res) -| (bilinear);
  \end{scope}
\end{tikzpicture}}}
\subfigure[Convolutional layers\label{fig:ms3_arch:conv_layers}]{
  \scalebox{0.33}{\begin{tikzpicture}[
  black!50,
  text=black!80,
  fnode/.style={
    align=center,
    rectangle,minimum height=40pt,minimum width=40pt,rounded corners=2pt,
    inner sep=5pt,
    line width=3pt,
    fill=pnodefill,draw=pnodedraw,
    font=\ttfamily\Large},
  dnode/.style={
    fnode,trapezium,trapezium left angle=70, trapezium right angle=110,trapezium stretches},
  inode/.style={
    dnode,fill=inodefill,draw=inodedraw},
  pnode/.style={
    fnode,fill=pnodefill,draw=pnodedraw},
  pnode2/.style={
    fnode,fill=pnodefill!50!white,draw=pnodedraw!50!white},
  onode/.style={
    dnode,fill=onodefill,draw=onodedraw},
  smnode/.style={
    mnode,fill=mnodefill,draw=mnodedraw},
  flow/.style={
    -latex,shorten >=1pt,line width=3pt,line cap=round,rounded corners=2pt,draw=pnodedraw,draw=#1},
  flow2/.style={
    flow,draw=pnodedraw!50!white}]
  \matrix[row sep=20pt,column sep=19pt,ampersand replacement=\&] {
	\node (conv1) [pnode]  {conv 5x5\\16 maps}; \& 
	\node (bn1)   [pnode]  {batch\\norm}; \&
	\node (relu1) [pnode]  {ReLU}; \&
	
	\node (conv2) [pnode2] {conv 3x3\\32 maps}; \& 
	\node (bn2)   [pnode2] {batch\\norm}; \&
	\node (relu2) [pnode2] {ReLU}; \&
	
	\node (conv3) [pnode]  {conv 3x3\\32 maps}; \& 
	\node (bn3)   [pnode]  {batch\\norm}; \&
	\node (relu3) [pnode]  {ReLU}; \&
	
	\node (conv4) [pnode2] {conv 1x1\\32 maps}; \& 
	\node (bn4)   [pnode2] {batch\\norm}; \&
	\node (relu4) [pnode2] {ReLU}; \&
	
	\node (conv5) [pnode]  {conv 1x1\\2 maps}; \& 
	\node (tanh5) [pnode]  {TanH}; \\
  };
  
  \begin{scope}[on background layer]
    \path (conv1) edge[flow] (bn1);
    \path (bn1) edge[flow] (relu1);
    \path (relu1) edge[flow] (conv2);
    
    \path (conv2) edge[flow2] (bn2);
    \path (bn2) edge[flow2] (relu2);
    \path (relu2) edge[flow2] (conv3);
    
    \path (conv3) edge[flow] (bn3);
    \path (bn3) edge[flow] (relu3);
    \path (relu3) edge[flow] (conv4);
    
    \path (conv4) edge[flow2] (bn4);
    \path (bn4) edge[flow2] (relu4);
    \path (relu4) edge[flow2] (conv5);
    
    \path (conv5) edge[flow] (tanh5);
  \end{scope}
\end{tikzpicture}}}
\subfigure[Legend\label{fig:ms3_arch:legend}]{
  \scalebox{0.33}{\begin{tikzpicture}[
  black!50,
  text=black!80,
  lnode/.style={
    minimum height=10pt,minimum width=80pt,font=\ttfamily\Large},
  fnode/.style={
    align=center,
    rectangle,minimum height=40pt,minimum width=80pt,rounded corners=2pt,
    inner sep=5pt,
    line width=3pt,
    fill=pnodefill,draw=pnodedraw,
    font=\ttfamily\Large},
  dnode/.style={
    fnode,trapezium,trapezium left angle=70, trapezium right angle=110},
  inode/.style={
    dnode,fill=inodefill,draw=inodedraw},
  pnode/.style={
    fnode,fill=pnodefill,draw=pnodedraw},
  onode/.style={
    dnode,fill=onodefill,draw=onodedraw},
  mnode/.style={
    pnode,isosceles triangle,isosceles triangle stretches,minimum width=35pt,minimum height=35pt},
  smnode/.style={
    mnode,fill=mnodefill,draw=mnodedraw},
  flow/.style={
    -latex,shorten >=1pt,line width=3pt,line cap=round,rounded corners=2pt,draw=pnodedraw,draw=#1},
  branch/.style={
    circle,inner sep=0pt,minimum size=8pt,fill=pnodedraw,draw=pnodedraw,fill=#1,draw=#1},
  vhedge/.style={
    rounded corners,to path=|- (\tikztotarget)}]
  \newcommand{\mytrap}[5]{%
    \node (#1)  [anchor=center,#2] {\phantom{#4}\\\phantom{#5}}; \node[anchor=center,font=\ttfamily\Large] at (#1.center) {#3};
  }
  \matrix[row sep=10pt,column sep=40pt,ampersand replacement=\&] {
  	\node (input_data)  [inode]                   {\phantom{input}\\\phantom{data}}; \&
  	\mytrap{data}{dnode}{}{input}{data} \&
  	\node (out)         [onode]                   {\phantom{output}\\\phantom{data}}; \&
    \node (proc_block)  [pnode]                   {\phantom{processing}\\\phantom{block}}; \& 
	\node (merge)       [mnode]                   {}; \\
	\node[lnode] {input data}; \&
	\node[lnode] {data}; \&
	\node[lnode] {output data}; \&
	\node[lnode] {processing block}; \&
	\node[lnode] {concatenate}; \\
  };
\end{tikzpicture}}}
\caption{The {\bf basic warping architecture} \ref{fig:ms3_arch:high_level} takes an input eye region augmented with eye feature points information (\texttt{input}) as well as a correction \texttt{angle} and produces an image of the redirected eye. The model contains three main blocks: angle embedding module (\texttt{embed angle}) calculating a vector representation of the correction \texttt{angle} and two warping modules (\texttt{process $ 0.5\times $-scale} \ref{fig:ms3_arch:05_scale} and \texttt{process $ 1\times $-scale} \ref{fig:ms3_arch:1_scale}) predicting and applying pixel-flow to the input image.}
\label{fig:ms3_arch}
\end{figure*}
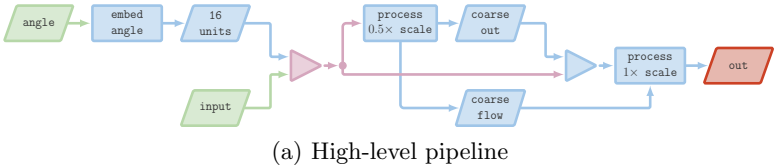
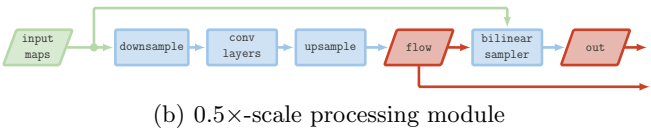
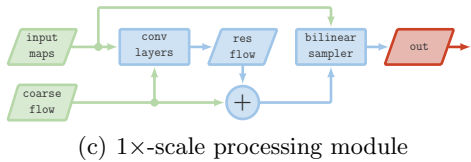
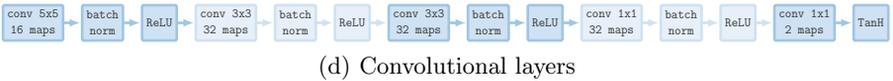
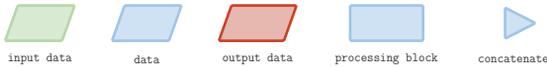

\end{appendices}

\end{document}